\documentclass[lettersize,journal]{IEEEtran}
\usepackage{amsmath,amsfonts}
\usepackage{algorithmic}
\usepackage{array}
\usepackage[caption=false,font=normalsize,labelfont=sf,textfont=sf]{subfig}
\usepackage{textcomp}
\usepackage{stfloats}
\usepackage{url}
\usepackage{verbatim}
\usepackage{graphicx}
\def\BibTeX{{\rm B\kern-.05em{\sc i\kern-.025em b}\kern-.08em
    T\kern-.1667em\lower.7ex\hbox{E}\kern-.125emX}}
\usepackage{balance}
\usepackage[dvipsnames]{xcolor} % enables extended color names including ForestGreen
\usepackage{tikz}
\usepackage{booktabs}
\usepackage{graphicx} % For resizebox
\usepackage{amsmath}  % For math symbols if needed elsewhere
\usepackage{array}    % For column definitions if needed
\usepackage{multirow} % Needed for multirow
\usepackage{mathtools}
\usepackage{multirow}
\usepackage{booktabs}

\newcommand*\circled[1]{\tikz[baseline=(char.base)]{
    \node[shape=circle,fill=black,inner sep=0.3pt] (char) {\textcolor{white}{#1}};}}

\newcommand*\circledgreen[1]{%
  \tikz[baseline=(char.base)]{%
    \node[shape=circle, fill=ForestGreen, inner sep=0.3pt] (char) {\textcolor{white}{#1}};%
  }%
}

\begin{document}
\title{MARBLE: A Multi-Agent Rule-Based LLM Reasoning Engine for Accident Severity Prediction}

\author{Kaleem Ullah Qasim, Jiashu Zhang%
\thanks{Kaleem Ullah Qasim and Jiashu Zhang are with the School of Computing and Artificial Intelligence, Southwest Jiaotong University, Chengdu, Sichuan, China (e-mail: kaleem@my.swjtu.edu.cn; jszhang@home.swjtu.edu.cn).}%
\thanks{This work was supported by the National Natural Science Foundation of China under Grant 62471411}%
\thanks{Corresponding author: Jiashu Zhang.}
}

\markboth{Journal of \LaTeX\ Class Files,~Vol.~18, No.~9, September~2020}%
{How to Use the IEEEtran \LaTeX \ Templates}

\maketitle

\begin{abstract}
Accident severity prediction plays a critical role in transportation safety systems but is a persistently difficult task due to incomplete data, strong feature dependencies, and severe class imbalance in which rare but high-severity cases are underrepresented and hard to detect. Existing methods often rely on monolithic models or black-box prompting, which struggle to scale in noisy, real-world settings and offer limited interpretability. To address these challenges, we propose MARBLE: a multi-agent rule-based LLM engine that decomposes the severity prediction task across a team of specialized reasoning agents, including an interchangeable ML-backed agent. Each agent focuses on a semantic subset of features (e.g., spatial, environmental, temporal), enabling scoped reasoning and modular prompting without the risk of prompt saturation. Predictions are coordinated through either rule-based or LLM-guided consensus mechanisms that account for class rarity and confidence dynamics. The system retains structured traces of agent-level reasoning and coordination outcomes, supporting in-depth interpretability and post-hoc performance diagnostics. Across both UK and US datasets, MARBLE consistently outperforms traditional machine learning classifiers and state-of-the-art (SOTA) prompt-based reasoning methods—including Chain-of-Thought (CoT), Least-to-Most (L2M), and Tree-of-Thought (ToT)—achieving nearly 90\% accuracy where others plateau below 48\%. This performance redefines the practical ceiling for accident severity classification under real-world noise and extreme class imbalance. Our results position MARBLE as a generalizable and interpretable framework for reasoning under uncertainty in safety-critical applications.
\end{abstract}


\section{Introduction}
\label{sec:introduction}

Traffic accidents impose staggering societal costs, demanding effective strategies for mitigation and emergency response \cite{dimitriou2016cost,Li2023CrashSeverity,Vinta2024BConvLSTM}. Central to enhancing road safety and optimizing resource allocation is the accurate prediction of accident severity. This high-stakes task involves forecasting the injury outcome—ranging from minor incidents to fatalities—based on a diverse set of contributing factors captured at the scene \cite{yang2024crash,gao2023uncertainty}. However, predicting severity remains a persistent challenge due to the inherent complexities of real-world accident data, which is often characterized by incompleteness, strong non-linear feature dependencies, and severe class imbalance where critical, high-severity events are significantly underrepresented \cite{jiang2025analyzing,yang2024crash,Ghosh2024ClassImbalance,Borisov2024TabularSurvey}.

While classical machine learning and more recently deep learning models have shown promise in mapping structured accident features (e.g., weather, road type, time) to severity classes \cite{Pei28052025,li2024interpretable}, they often falter under these real-world conditions \cite{Manzoor2021RFCNN,Arrieta2020XAI}. Traditional ML models struggle to capture complex interactions between diverse feature types, leading to potential overfitting on majority classes (typically less severe accidents) and brittle decision boundaries when faced with noisy or incomplete inputs. Deep learning models, while more expressive, demand substantial labeled training data often unavailable for rare event prediction and typically operate as "black boxes." This lack of transparency hinders trust and diagnostic analysis, which are crucial in safety-critical applications. Furthermore, both paradigms often implicitly assume feature independence or uniform relevance, an assumption frequently violated in accident scenarios where factors such as weather and road geometry interact in complex ways.

The advent of transformer-based large language models (LLMs) has introduced advanced reasoning capabilities, exemplified by techniques like Chain-of-Thought (CoT) prompting \cite{Wei2022CoT,Wang2023SelfConsistency}. Concurrently, the field of agentic AI is rapidly exploring how autonomous, goal-directed LLM agents can decompose and solve complex problems \cite{Pateria2021HRLsurvey,Shen2023HuggingGPT}. However, applying these concepts directly to structured prediction tasks like accident severity classification faces significant hurdles \cite{Li2024LLM_MAS}. Full-scale LLMs, the foundation for most current agent frameworks, remain computationally expensive, with high inference latency and memory footprints unsuitable for real-time deployment. More fundamentally, directing a single, general-purpose LLM to reason over high-dimensional, tabular data often leads to overloaded prompts where dozens of features entangle the reasoning process. This diminishes accuracy and obscures the identification of salient causal factors essential for interpretability, contrasting sharply with the potential of specialized, cooperative multi-agent systems.

These limitations highlight a critical gap: the need for a framework that leverages the decompositional power \cite{qasim2025recursivedecompositionlogicalthoughts} of agentic AI but is adapted specifically for structured data prediction under resource constraints. We posit that a paradigm shift is necessary towards distributed systems of specialized reasoning agents grounded in efficient models. By decomposing the complex problem space and assigning semantically coherent subsets of features to dedicated, lightweight agents focused on specific domains (e.g., environmental, temporal), it may be possible to achieve high-fidelity predictions and interpretability without the drawbacks of monolithic approaches.

\begin{figure*}[htbp]
    \centering
    \includegraphics[width=1\linewidth]{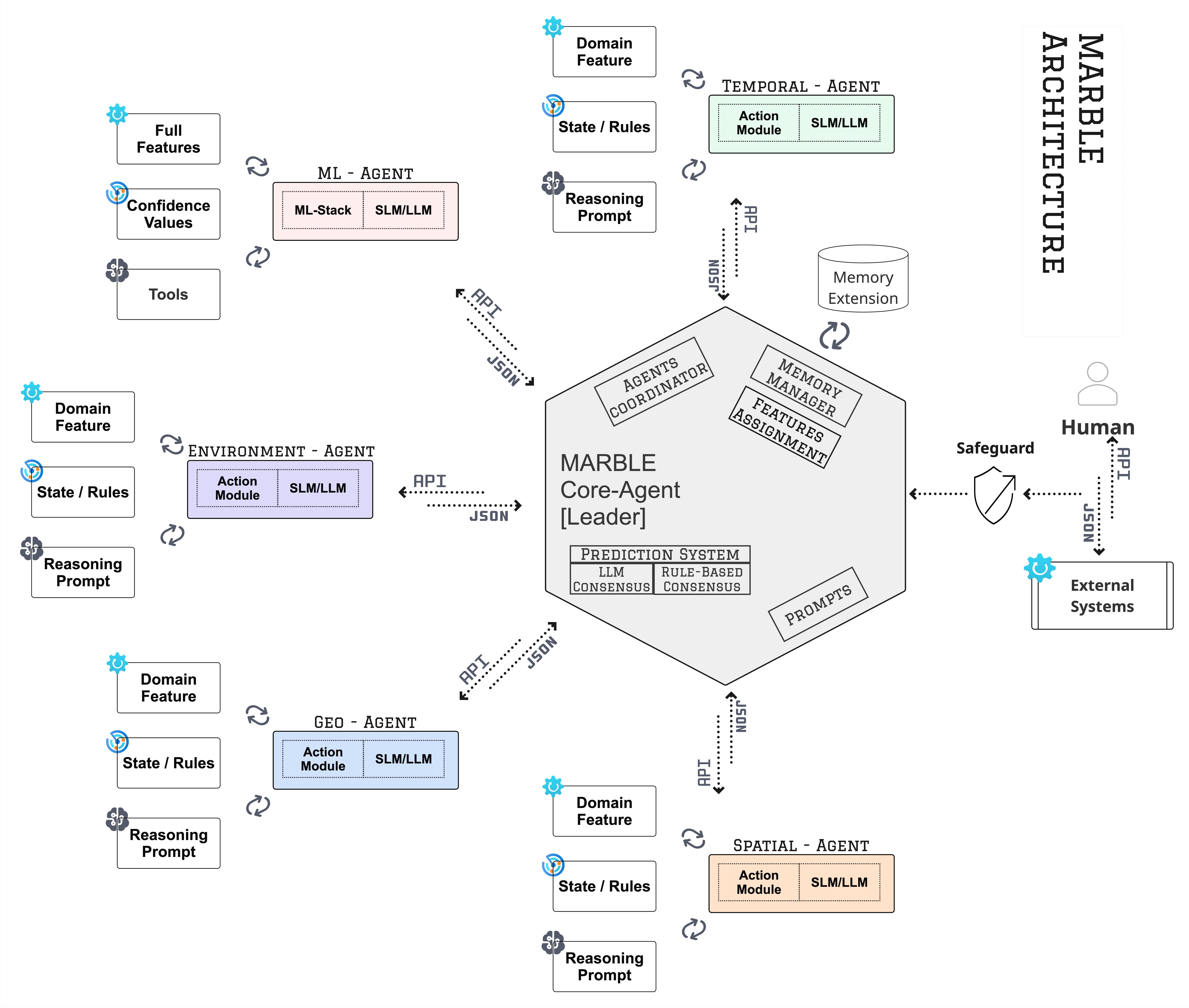}
    \caption{Overview of the MARBLE architecture with a Core-Agent coordinating domain-specific agents and consensus-based inference.}
    \label{fig:marble-architecture}
\end{figure*}

To address these challenges, we introduce MARBLE: a Multi-Agent Rule-Based LLM Reasoning Engine for accident severity prediction. MARBLE operationalizes cooperative multi-agent reasoning by decomposing inference across a distributed system of feature-specialized agents. An interchangeable ML-backed agent provides a baseline prediction, while multiple domain-specific agents, powered by computationally efficient Small Language Models (SLMs), analyze distinct semantic dimensions such as environmental, spatial, and temporal factors \cite{Magister2023SmallLMReason,Li2023SCoTD,Zhu2024MathDistill}. This modular architecture, inspired by agentic decomposition, directly mitigates the context saturation of monolithic models by providing each agent with only its relevant feature subset. The use of SLMs ensures that MARBLE retains the powerful inference capabilities of transformers while enabling practical deployment.

Predictions from this ensemble of diverse agents are synthesized by a central Coordinator. This is achieved using either interpretable rule-based logic or, optionally, LLM-guided reflection, effectively implementing a structured coordination protocol for the agent team \cite{Papoudakis2021MarlBenchmark,Ning2024MARLsurvey}. To further enhance reliability without extensive model retraining, advanced prompting techniques like CoT and Least-to-Most (L2M) are strategically embedded at both the agent level (for focused reasoning) and the coordination level (for structured synthesis) \cite{Zhou2023LeastToMost,Kojima2022ZeroShotCoT}. The system maintains structured traces of all agent-level reasoning and coordination outcomes, fostering transparency and aligning with the goals of explainable agentic systems.

To the best of our knowledge, MARBLE is the first framework to instantiate a hybrid multi-agent system (combining ML and domain-specific SLM agents) tailored for structured tabular data prediction. It employs lightweight SLM-based agentic reasoning, multi-level prompting, and structured coordination to tackle accident severity prediction under severe class imbalance \cite{Park2023GenerativeAgents}. Our contributions are as follows:

\noindent
\hangindent=1.5em \textbf{\circledgreen{1}} We propose MARBLE, a novel multi-agent hybrid reasoning system synergizing structured ML prediction with prompt-guided SLM coordination, demonstrating an effective application of agentic principles to robust accident severity classification.

\noindent
\hangindent=1.5em \textbf{\circledgreen{2}} We introduce a modular agent architecture that partitions the feature space across domain-specialized agents, enabling decomposable, interpretable reasoning while mitigating context overload and enhancing focus.

\noindent
\hangindent=1.5em \textbf{\circledgreen{3}} We show empirically that MARBLE achieves state-of-the-art performance on benchmark accident severity datasets, significantly outperforming traditional ML/DL baselines and advanced prompting methods, particularly under data scarcity and class imbalance, while maintaining practical computational efficiency.

\section{Problem Statement}
\label{sec:problem_statement}
The task is to predict traffic accident severity $y \in \mathcal{Y} = \{1, ..., K\}$ from a feature vector $\mathbf{x} \in \mathcal{F}$. This prediction is challenged by severe \textit{class imbalance}, complex \textit{feature interactions}, and the need for high interpretability in this safety-critical context.

We frame this task within a \textit{Multi-Agent System (MAS)}, decomposing the overall prediction $F: \mathcal{F} \rightarrow \mathcal{Y}$ across a set of specialized agents $\mathcal{A} = \{a_1, ..., a_N\}$. This set includes a machine learning agent ($a_{ML}$) and several domain-specific SLM agents ($a \in \mathcal{A}_{SLM}$). Each agent $a_i$ processes a feature subset $\mathbf{x}_i = \pi_i(\mathbf{x})$ using its reasoning function $f_i: \mathcal{F}_i \rightarrow \mathcal{Y} \times [0, 1] \times \mathcal{R}_i$. This yields a structured output $O_i = (\hat{y}_i, c_i, r_i)$, which comprises the agent's prediction, confidence, and reasoning trace. The central problem is thus to design an effective \textit{coordination mechanism} $\Phi$ that synthesizes the collective agent outputs $\{O_i\}_{i=1}^N$ into a final prediction $\hat{Y}$ and confidence $C_{final}$. Formally, $\Phi: \{O_i\} \rightarrow (\hat{Y}, C_{final})$. An effective coordinator must leverage agent confidences ($c_i$), manage disagreements, address class imbalance, and maintain system-level interpretability. The following section details MARBLE's instantiation of the agent functions ($f_i$), feature projections ($\pi_i$), and coordination strategy ($\Phi$) to solve this problem.

\section{Methodology}
\label{sec:methodology}

The MARBLE framework formalizes accident severity prediction as a multi-stage process. The methodology begins with a feature engineering pipeline to construct the input space $\mathcal{F}$. This engineered input is then processed in parallel by a heterogeneous set of agents, $\mathcal{A}$, each with a distinct reasoning function: a primary machine learning agent ($f_{ML}$) that operates on the full feature set, and several domain-specialized SLM agents ($f_{a}$) that reason over specific feature subsets ($\pi_a(\mathbf{x})$). The outputs from all agents are synthesized by a coordination mechanism, for which we define both a deterministic rule-based logic ($\Phi_{RB}$) and an alternative LLM-based approach ($\Phi_{LLM}$). Finally, a decision selection function ($F_{final}$) integrates these signals to produce the ultimate severity prediction $\hat{Y}$ and system confidence $C_{final}$. Each component is detailed in the following subsections.

\subsection{System Architecture and Design Definitions}
\label{subsec:system_architecture}

MARBLE addresses the supervised prediction of accident severity, $y \in \mathcal{Y}$, from a feature vector $\mathbf{x} \in \mathcal{F} \subseteq \mathbb{R}^d$. The severity classes are defined as $\mathcal{Y} = \{1, 2, 3, 4\}$, where the task is characterized by severe class imbalance. The objective is to derive a final prediction $\hat{Y} = \arg\max_{y \in \mathcal{Y}} P(y | \mathbf{x})$. The complete system is an ensemble function, $F_{MARBLE}$, which maps an input $\mathbf{x}$ to a final prediction tuple $(\hat{Y}, C_{final})$:
\begin{equation}
F_{MARBLE}: \mathcal{F} \rightarrow \mathcal{Y} \times [0, 1]
\label{eq:marble_function_signature}
\end{equation}

The architecture is an ensemble of heterogeneous agents, $\mathcal{A}$, comprising a primary machine learning model ($a_{ML}$) and several domain-specific SLM agents:
\begin{equation}
\mathcal{A} = \{a_{ML}, a_{Env}, a_{Loc}, a_{Spa}, a_{Temp}\}
\label{eq:agent_set}
\end{equation}
Each agent $a \in \mathcal{A}$ implements a function $f_a: \mathcal{X}_a \rightarrow \mathcal{Y} \times [0, 1] \times \mathcal{R}_a$, which maps an agent-specific input $\mathbf{x}_a$ to an output tuple containing its prediction, confidence, and reasoning. The collective agent outputs, $O = \{(\hat{y}_a, c_a, r_a) | a \in \mathcal{A}\}$, are processed by a coordination mechanism $\Phi$ to produce a coordinated prediction $(\hat{y}_{coord}, C_{coord})$. A final decision logic, $F_{final}$, then integrates the coordinator's output with the ML agent's direct output to determine the system's ultimate prediction, $(\hat{Y}, C_{final})$.

\subsection{Feature Engineering, Representation, and Distribution}
\label{subsec:feature_engineering}

A core principle of MARBLE is the strategic partitioning of this feature space $\mathcal{F}$. While the primary machine learning agent ($a_{ML}$) operates on the complete feature set, domain-specific SLM agents ($a \in \mathcal{A}_{SLM}$) receive lower-dimensional, semantically coherent inputs. This is achieved via agent-specific projection functions $\pi_a$:
\begin{equation}
\pi_a: \mathcal{F} \rightarrow \mathcal{F}_a, \quad \text{where } \mathcal{F}_a \subset \mathcal{F} \text{ for } a \in \mathcal{A}_{SLM}
\label{eq:projection_function}
\end{equation}
Each projection $\pi_a$ selects a feature subset relevant to an agent's expertise (e.g., environmental, spatial, temporal). A listing of these subsets (not limited) is provided in Appendix ~\ref{tab:features_list}.

This strategic distribution of features, where each SLM agent receives $\mathbf{x}_a = \pi_a(\mathbf{x})$, serves three critical purposes: \circled{1} it provides each SLM agent with a focused, semantically coherent context, enhancing its specialized local reasoning; \circled{2} it significantly reduces the input dimensionality for SLMs ($|\mathcal{F}_a| \ll d$), thereby minimizing token consumption and computational overhead during inference; and \circled{3} it improves system robustness by isolating agents from noise or irrelevant information present in features outside their specific domain.

\subsection{Agent Function Formalization ($f_a$)}
\label{subsec:agent_formalization}

Each agent function $f_a$ processes its designated input features and produces an output tuple $(\hat{y}_a, c_a, r_a)$, comprising a predicted severity class $\hat{y}_a \in \mathcal{Y}$, a calibrated confidence score $c_a \in [0, 1]$, and potentially associated reasoning $r_a$ represented in a suitable format $\mathcal{R}_a$ (e.g., textual explanation for SLMs, or null/feature importance vectors for ML).

\subsubsection{Machine Learning Agent ($f_{ML}$)}
The machine learning agent, $a_{ML}$, leverages a conventionally trained predictive model, denoted $M_{ML}$ (e.g., a RandomForest or Gradient Boosting classifier optimized on $D_{train}$), which operates on the complete engineered feature space $\mathcal{F}$. Given an input vector $\mathbf{x} \in \mathcal{F}$, the model $M_{ML}$ yields a probability distribution over the severity classes $\mathcal{Y}$. Let $p_{ML}(y=k | \mathbf{x})$ represent the estimated probability that the true class is $k$, given $\mathbf{x}$, for $k \in \mathcal{Y}$. The prediction $\hat{y}_{ML}$ is determined by selecting the class with the maximum posterior probability:
\begin{equation}
    \hat{y}_{ML} = \arg\max_{k \in \mathcal{Y}} p_{ML}(y=k | \mathbf{x})
    \label{eq:ml_prediction}
\end{equation}
The initial confidence score, $c'_{ML}$, is naturally defined as this maximum probability value:
\begin{equation}
    c'_{ML} = \max_{k \in \mathcal{Y}} p_{ML}(y=k | \mathbf{x})
    \label{eq:ml_raw_confidence}
\end{equation}
For the ML agent within the current MARBLE implementation, no additional calibration beyond the model's inherent probability estimation is applied, thus the final confidence is $c_{ML} = c'_{ML}$. The reasoning component for the ML agent, $r_{ML}$, is considered null ($r_{ML} = \emptyset$) in the context of the coordination input, although feature importance measures derived from $M_{ML}$ was computed for post-hoc analysis. The complete function for the ML agent is therefore defined as:
\begin{equation}
\begin{split}
f_{ML}(\mathbf{x}) = (\hat{y}_{ML}, c_{ML}, \emptyset) = \bigg( 
& \arg\max_{k \in \mathcal{Y}} p_{ML}(k | \mathbf{x}), \\
& \max_{k \in \mathcal{Y}} p_{ML}(k | \mathbf{x}), \emptyset 
\bigg)
\end{split}
\label{eq:ml_function}
\end{equation}

\subsubsection{Small Language Model Agents ($f_a$ for $a \in \mathcal{A}_{SLM}$)}
The domain-specific agents $a \in \mathcal{A}_{SLM} = \mathcal{A} \setminus \{a_{ML}\}$ utilize pre-trained SLMs to perform reasoning over their assigned feature subsets $\mathcal{F}_a$. For a given system input $\mathbf{x} \in \mathcal{F}$, the corresponding agent input is the projected feature vector $\mathbf{x}_a = \pi_a(\mathbf{x}) \in \mathcal{F}_a$. This vector $\mathbf{x}_a$ undergoes an implicit preprocessing step $\Psi_a: \mathcal{F}_a \rightarrow \mathcal{X}_a$, primarily involving the formatting of numerical and categorical features into a structured textual representation suitable for the SLM prompt.

The core of the SLM agent's operation involves dynamic prompt generation via a function $P_a: \mathcal{X}_a \rightarrow \text{String}$. This function constructs the input prompt $prompt_a$ by assembling predefined template components with the formatted feature information:

\begin{equation}
\begin{split}
prompt_a = P_a(\Psi_a(\mathbf{x}_a)) =\  
& T_{\text{context}, a} \oplus T_{\text{instr}, a} \\
& \oplus\ \text{Format}(\Psi_a(\mathbf{x}_a)) \oplus T_{\text{query}, a}
\end{split}
\label{eq:prompt_generation}
\end{equation}

Here, $\oplus$ denotes string concatenation, $T_{context, a}$ provides domain background, $T_{instr, a}$ gives specific instructions, $\text{Format}(\cdot)$ serializes the input features $\Psi_a(\mathbf{x}_a)$, and $T_{query, a}$ poses the specific question regarding severity prediction and confidence assessment. For instance, the $\text{Format}(\Psi_a(\mathbf{x}_a))$ might resemble "Weather: Rainy, Visibility: 0.5 miles, ...".

The generated $prompt_a$ is then processed by the designated SLM (a pre-trained transformer model), yielding a textual output $o_a$:
\begin{equation}
    o_a = \text{SLM}(prompt_a)
    \label{eq:slm_invocation}
\end{equation}
To extract the required structured information from the free-form text output $o_a$, an extraction function $E_a: \text{String} \rightarrow \mathcal{Y} \times [0,1] \times \text{String}$ is applied. This function typically employs regular expressions or parsing of a requested format (e.g., JSON embedded in the text) to retrieve the agent's predicted severity class $\hat{y}_a$, its raw confidence estimate $c'_a \in [0,1]$ (as stated by the SLM in its response), and the textual reasoning $r_a$ provided by the SLM:
\begin{equation}
    (\hat{y}_a, c'_a, r_a) = E_a(o_a)
    \label{eq:structured_extraction}
\end{equation}
The raw confidence score $c'_a$ obtained directly from the SLM output may not be reliably calibrated. Therefore, we apply an agent-specific calibration function $Calibrate_a: [0,1] \times \mathcal{Y} \rightarrow [0,1]$, derived empirically to obtain the final calibrated confidence $c_a$. This function applies a heuristic boost to high-confidence predictions for the underrepresented rare classes ($y \in \{1, 4\}$):

\begin{equation}
\begin{aligned}
c_a &= \text{Calibrate}_a(c'_a, \hat{y}_a) = \\
&\begin{cases}
\min(0.98,\ c'_a + 0.1) & \text{if } \hat{y}_a \in \{1,4\},\ c'_a > 0.8 \\
\min(0.9,\ c'_a + 0.05) & \text{if } \hat{y}_a \in \{1,4\},\ c'_a > 0.6 \\
c'_a & \text{otherwise}
\end{cases}
\end{aligned}
\label{eq:confidence_calibration}
\end{equation}

The complete function for an SLM agent $a \in \mathcal{A}_{SLM}$ encapsulates these steps:
\begin{equation}
    f_a(\mathbf{x}) = (\hat{y}_a, c_a, r_a) = (E_a \circ \text{SLM} \circ P_a \circ \Psi_a \circ \pi_a)(\mathbf{x})
    \label{eq:slm_function}
\end{equation}
where $\circ$ denotes function composition. This formulation highlights the distinct processing pipeline for SLM agents compared to the ML agent.

\subsection{Agents Communication Protocol}
MARBLE’s communication protocol is centralized and orchestrated by a Coordinator agent (Fig.~\ref{fig:marble-architecture}). The inference cycle for an instance $\mathbf{x}$ follows a synchronous, three-stage process: \circled{1} The Coordinator receives the full feature vector $\mathbf{x}$. It then applies a predefined projection function $\pi_i$ for each agent $\mathcal{A}_i$ to generate a tailored feature subset $\mathbf{x}_i = \pi_i(\mathbf{x})$. Each subset $\mathbf{x}_i$ is then dispatched to its corresponding agent. \circled{2} Upon receiving its specific feature subset $\mathbf{x}_i$, each agent performs its local reasoning (via its ML model or SLM prompt) and returns a structured tuple $\mathcal{R}_i = (s_i, c_i, r_i)$, comprising its prediction, confidence, and reasoning trace.\circled{3} The Coordinator, which is blocking, waits to receive a response from all agents. It then aggregates their individual outputs into a complete set for downstream processing:

\begin{equation}
\mathcal{M} = \left\{ \mathcal{R}_i \mid i = 1, \dots, N \right\}
\label{eq:memory_buffer}
\end{equation}
The resulting set $\mathcal{M}$ serves as the complete input for the coordination mechanism. This protocol treats agents as stateless functions that operate independently on the specific data they receive, with no direct inter-agent communication.

\begin{table*}[ht]
    \centering
    \caption{Performance Comparison: MARBLE (Best) vs. Baselines and Prompting Techniques}
    \renewcommand{\arraystretch}{1.4}
    \setlength{\tabcolsep}{3.5pt} 
    \small 
    \resizebox{\textwidth}{!}{% 
    \begin{tabular}{@{}lcccccccc@{}} 
        \toprule
        \multirow{2}{*}{\textbf{Methods}} & \multicolumn{4}{c}{\textbf{UK Dataset}} & \multicolumn{4}{c}{\textbf{US Dataset}} \\
        \cmidrule(lr){2-5} \cmidrule(lr){6-9}
        & \textbf{Accuracy (\%)} & \textbf{Precision} & \textbf{Recall} & \textbf{F1 Score} & \textbf{Accuracy (\%)} & \textbf{Precision} & \textbf{Recall} & \textbf{F1 Score} \\
        \midrule
        \multicolumn{9}{c}{\textbf{Machine Learning Models (Baselines)}} \\
        \midrule
        Random Forest & 41.2 & 0.38 & 0.42 & 0.39 & 43.5 & 0.40 & 0.44 & 0.41 \\
        Gradient Boosting & 42.8 & 0.39 & 0.44 & 0.39 & 44.7 & 0.41 & 0.45 & 0.42 \\
        Support Vector Machine & 40.1 & 0.37 & 0.41 & 0.38 & 42.0 & 0.39 & 0.43 & 0.40 \\
        LSTM & 45.6 & 0.42 & 0.46 & 0.44 & 47.3 & 0.44 & 0.48 & 0.46 \\
        \midrule
        \multicolumn{9}{c}{\textbf{Prompting Systems (Executed on LLaMA 3.2 3B)}} \\
        \midrule
        Vanilla & 25.8 & 0.23 & 0.24 & 0.23 & 26.2 & 0.24 & 0.25 & 0.24 \\
        Chain-of-Thought \cite{chain_of_thought_prompting_elicits_reasoning_in_large_language_models} & 27.6 & 0.25 & 0.26 & 0.25 & 28.8 & 0.26 & 0.27 & 0.26 \\
        Self-Consistency \cite{self_consistency_improves_chain_of_thought_reasoning_in_language_models}& 29.1 & 0.27 & 0.28 & 0.27 & 30.2 & 0.28 & 0.29 & 0.28 \\
        Least-to-Most \cite{least_to_most_prompting_enables_complex_reasoning_in_large_language_models} & 24.5 & 0.22 & 0.23 & 0.22 & 25.4 & 0.23 & 0.24 & 0.23 \\
        Graph-of-Thought \cite{graph_of_thoughts_solving_elaborate_problems_with_large_language_models} & 30.2 & 0.28 & 0.29 & 0.28 & 31.5 & 0.29 & 0.30 & 0.29 \\
        Tree-of-Thought \cite{tree_of_thoughts_deliberate_problem_solving_with_large_language_models} & 30.5 & 0.29 & 0.30 & 0.29 & 31.8 & 0.30 & 0.31 & 0.30 \\
        Chain-of-Draft \cite{CoD} & 28.9 & 0.26 & 0.27 & 0.26 & 29.1 & 0.27 & 0.28 & 0.27 \\
        \midrule
        \multicolumn{9}{c}{\textbf{MARBLE (Executed on Smollm2 1.7B)}} \\
        \midrule
         MARBLE (Best)* & \textbf{89.5}$\uparrow$ & \textbf{0.91}$\uparrow$ & \textbf{0.89}$\uparrow$ & \textbf{0.90}$\uparrow$ & \textbf{89.8}$\uparrow$ & \textbf{0.90}$\uparrow$ & \textbf{0.89}$\uparrow$ & \textbf{0.89}$\uparrow$ \\ 
        \bottomrule
    \end{tabular}%
    } % End resizebox
    \label{tab:marble_vs_baselines}
    \\ % Add line break for the note
    \footnotesize{\textit{Note:} To provide a robust performance evaluation and minimize dependency on any single data split, the reported results are the averaged metrics from 5 independent runs in a cross-validation framework.}
\end{table*}

\subsection{Coordination Mechanisms}
\label{subsec:coordination}

Following the execution and output generation by the individual agents $a \in \mathcal{A}$, as formalized in Sec.~\ref{subsec:agent_formalization}, the MARBLE system employs a coordination mechanism $\Phi$ to synthesize these potentially diverse perspectives into a unified prediction. The input to this stage is the set of agent output tuples:
\begin{equation}
    O = \{(\hat{y}_a, c_a, r_a) | a \in \mathcal{A}\}
    \label{eq:agent_outputs_set}
\end{equation}
MARBLE incorporates two distinct coordination strategies: a deterministic Rule-Based Coordination mechanism ($\Phi_{RB}$), which serves as the default, and an alternative LLM-Based Coordination mechanism ($\Phi_{LLM}$) which was inspired by recent research \cite{gu2025surveyllmasajudge}. Both mechanisms aim to produce a coordinated severity prediction $\hat{y}_{coord} \in \mathcal{Y}$ and an associated confidence score $C_{coord} \in [0, 1]$.

\subsubsection{Rule-Based Coordination ($\Phi_{RB}$)}
\label{subsubsec:rb_coordination}
The rule-based mechanism $\Phi_{RB}: \mathcal{P}(\mathcal{Y} \times [0,1] \times \mathcal{R}) \rightarrow \mathcal{Y} \times [0,1]$ integrates agent outputs using a predefined set of heuristics, weights, and thresholds derived empirically from validation data ($D_{val}$). Central to this mechanism is a weighted voting scheme combined with specific override logic for high-confidence ML predictions.

First, we define the static agent importance weights, reflecting the perceived reliability or contribution of each agent, represented by the vector $\mathbf{w}$:

\begin{equation}
\begin{aligned}
\mathbf{w} &= (w_{ML}, w_{Env}, w_{Loc}, \\
            &\quad w_{Spa}, w_{Temp}) = (3.0,\ 1.5,\ 1.2,\ 1.0,\ 1.0)
\end{aligned}
\label{eq:agent_weights}
\end{equation}

Second, to explicitly address class imbalance, particularly for the rare classes $y \in \{1, 4\}$, we introduce class importance factors $\boldsymbol{\beta}$:
\begin{equation}
    \beta_k = \begin{cases} 1.2 & \text{if } k \in \{1, 4\} \\ 1.0 & \text{if } k \in \{2, 3\} \end{cases}
    \label{eq:class_factors}
\end{equation}
Using these parameters, a weighted voting score $S'_k$ is computed for each potential severity class $k \in \mathcal{Y}$ based on the input set $O$:
\begin{equation}
    S'_k(O, \mathbf{w}, \boldsymbol{\beta}) = \sum_{a \in \mathcal{A}} w_a \cdot c_a \cdot \beta_k \cdot \mathbb{I}(\hat{y}_a = k)
    \label{eq:weighted_vote_score}
\end{equation}
where $\mathbb{I}(\cdot)$ is the indicator function evaluates to 1 if the condition is true, and 0 otherwise.

The coordination logic incorporates an override mechanism prioritizing high-confidence predictions from the ML agent $a_{ML}$, subject to specific conditions. Let $\tau_{ML\_high} = 0.75$ and $\tau_{ML\_corrob} = 0.8$ be predefined confidence thresholds. Let $A_{SLM,k}(O) = \{a \in \mathcal{A}_{SLM} | \hat{y}_a = k\}$ denote the set of SLM agents predicting class $k$. The rule-based prediction $\hat{y}_{RB}$ is then determined as follows:

\begin{equation}
\hat{y}_{RB}(O) = \arg\max_{k} S'_k(O)
\label{eq:rb_prediction_stacked}
\end{equation}

In the event that the $\arg\max$ operation yields multiple classes with the identical maximal score $S'_{k_{max}}$, a specific tie-breaking rule is applied. Let $K_{max} = \{k \in \mathcal{Y} | S'_k(O) = \max_{j \in \mathcal{Y}} S'_j(O)\}$. If $|K_{max}| > 1$, the prediction favors the class supported by the minimum number of agents among the tied classes:

\begin{equation}
\begin{aligned}
\hat{y}_{RB}(O) &= \arg\min_{k \in K_{max}} |\{a \in \mathcal{A} \mid \hat{y}_a = k\}|, \\
&\quad \text{if } |K_{max}| > 1 \land\ \text{not override}
\end{aligned}
\label{eq:rb_tie_breaking}
\end{equation}

This heuristic aims to capture potentially unique signals in ambiguous cases. To refine the confidence assessment, an agreement-based boost $\Delta C$ is calculated based on the consensus among the SLM agents regarding the chosen prediction $\hat{y}_{RB}$. Let $A_k(O) = \{a \in \mathcal{A} | \hat{y}_a = k\}$ be the set of all agents predicting class $k$. The boost $\Delta C(O)$ is defined as:
\begin{equation}
\Delta C(O) =
\begin{cases}
  0.1, & \begin{aligned}[t]
            & \text{if } \hat{y}_{RB} \in \{1,4\} \text{ and} \\
            & |A_{SLM, \hat{y}_{RB}}| \geq 2 \text{ and} \\
            & \frac{|A_{SLM, \hat{y}_{RB}}|}{|\mathcal{A}_{SLM}|} > 0.5
         \end{aligned} \\[2ex] % Adds extra vertical space
  0.05, & \begin{aligned}[t]
             & \text{if } \hat{y}_{RB} \in \{2,3\} \text{ and} \\
             & \frac{|A_{SLM, \hat{y}_{RB}}|}{|\mathcal{A}_{SLM}|} > 0.5
          \end{aligned} \\
  0, & \text{otherwise}
\end{cases}
\label{eq:agreement_boost_stacked}
\end{equation}

Finally, the overall confidence $C_{RB}$ associated with the rule-based prediction $\hat{y}_{RB}$ is computed. If $\hat{y}_{RB}$ resulted from the ML override condition and t confidence is derived from $c_{ML}$. Otherwise, it is calculated using the weighted average confidence of the agreeing agents, denoted $\bar{c}_{aggr}(\hat{y})$ and defined as:
\begin{equation}
    \bar{c}_{aggr}(\hat{y}) = \frac{\sum_{a \in A_{\hat{y}}(O)} w_a c_a}{\sum_{a \in A_{\hat{y}}(O)} w_a}
    \label{eq:weighted_avg_conf}
\end{equation}
Using this weighted average, the final rule-based confidence $C_{RB}(O)$ is augmented by the agreement boost $\Delta C(O)$ and capped at 0.95, according to the following logic:
\begin{equation}
C_{RB}(O) =
\begin{cases}
  c_{ML}, & \begin{aligned}[t]
               & \text{if override and} \\
               & \hat{y}_{ML} \in \{2,3\}
            \end{aligned} \\[1ex]
  \min(0.95, c_{ML} + 0.15), & \begin{aligned}[t]
                                & \text{if override and} \\
                                & \hat{y}_{ML} \in \{1,4\}
                             \end{aligned} \\[1ex]
  \min(0.95, \bar{c}_{aggr}(\hat{y}_{RB}) + \Delta C), & \text{otherwise}
\end{cases}
\label{eq:rb_confidence_stacked}
\end{equation}
If the denominator $\sum_{a \in A_{\hat{y}_{RB}}(O)} w_a$ is zero (i.e., no agents predicted $\hat{y}_{RB}$, which should only occur in edge cases or error states), $C_{RB}$ is assigned a default low value (e.g., 0.1).

\begin{table*}[ht]
    \centering
    \caption{Performance Comparison of MARBLE Coordination Strategies}
    \renewcommand{\arraystretch}{1.4} 
    \setlength{\tabcolsep}{3.5pt} 
    \small 
    \resizebox{\textwidth}{!}{% 
    % Original column specifier without the pipe
    \begin{tabular}{@{}lcccccccc@{}} 
        \toprule
        % Changed the first column header to reflect what varies in this specific table
        \multirow{2}{*}{\textbf{Underlying SLM}} & \multicolumn{4}{c}{\textbf{UK Dataset}} & \multicolumn{4}{c}{\textbf{US Dataset}} \\ 
        \cmidrule(lr){2-5} \cmidrule(lr){6-9}
        & \textbf{Accuracy (\%)} & \textbf{Precision} & \textbf{Recall} & \textbf{F1 Score} & \textbf{Accuracy (\%)} & \textbf{Precision} & \textbf{Recall} & \textbf{F1 Score} \\
        \midrule
        \multicolumn{9}{c}{\textbf{MARBLE (LLM-Based Coordination)}} \\
        \midrule
        Gemma-3 1B & 65.8 & 0.62 & 0.64 & 0.63 & 67.5 & 0.64 & 0.66 & 0.65 \\
        LLaMA 3.2 1B & 69.2 & 0.66 & 0.68 & 0.67 & 70.9 & 0.68 & 0.70 & 0.69 \\
        Mistral 1.3B & 71.5 & 0.69 & 0.70 & 0.69 & 72.8 & 0.70 & 0.71 & 0.70 \\
        % Added arrows to indicate best performance WITHIN LLM-Based section
        LLaMA 3.2 3B & 72.1$\uparrow$ & 0.70$\uparrow$ & 0.71$\uparrow$ & 0.70$\uparrow$ & 73.4$\uparrow$ & 0.71$\uparrow$ & 0.72$\uparrow$ & 0.71$\uparrow$ \\ 
        \midrule
         \multicolumn{9}{c}{\textbf{MARBLE (Rule-Based Coordination)}} \\
        \midrule
        Gemma-3 1B & 79.5 & 0.76 & 0.78 & 0.77 & 81.2 & 0.78 & 0.80 & 0.79 \\
        LLaMA 3.2 1B & 83.1 & 0.80 & 0.82 & 0.81 & 84.8 & 0.82 & 0.84 & 0.83 \\
        Mistral 1.3B & 86.8 & 0.85 & 0.86 & 0.85 & 87.5 & 0.85 & 0.87 & 0.86 \\ % Removed previous best markers
        LLaMA 3.2 3B & 87.2 & 0.86 & 0.87 & 0.86 & 87.6 & 0.86 & 0.86 & 0.86 \\ % Removed previous best markers
        % Added new row with hypothetical US data. Bold+arrow only on OVERALL best values per column.
        Smollm2-1.7b* & \textbf{89.5}$\uparrow$ & \textbf{0.91}$\uparrow$ & \textbf{0.89}$\uparrow$ & \textbf{0.90}$\uparrow$ & \textbf{89.8}$\uparrow$ & \textbf{0.90}$\uparrow$ & \textbf{0.89}$\uparrow$ & \textbf{0.89}$\uparrow$ \\ 
        \bottomrule
    \end{tabular}%
    } % End resizebox
    \label{tab:marble_coordination_comparison_final_no_pipe} % Updated label
    \\ % Add line break for the note
    \footnotesize{\textit{HuggingFace's} SmollM2-1.7B was fine-tuned for agent-specific usage. It demonstrates robust performance when applied to domain-specialized agents within the MARBLE framework.}
\end{table*}

\subsubsection{LLM-Based Coordination ($\Phi_{LLM}$)}
\label{subsubsec:llm_coordination}
As an alternative, MARBLE supports coordination via a dedicated coordination SLM, denoted $\text{SLM}_{coord}$. This mechanism, $\Phi_{LLM}: \mathcal{P}(\mathcal{Y} \times [0,1] \times \mathcal{R}) \rightarrow \mathcal{Y} \times [0,1] \times \mathcal{R}$, employs meta-reasoning over the collected agent outputs $O$. A formatting function, $\text{Format}(O, \mathbf{w})$, serializes the agent predictions $\hat{y}_a$, confidences $c_a$, reasoning $r_a$, and potentially the agent weights $w_a$ into a structured meta-prompt $prompt_\Phi$. This prompt instructs $\text{SLM}_{coord}$ to synthesize the information and provide a final prediction, confidence, and summary reasoning.
\begin{equation}
    prompt_\Phi = \text{Format}(O, \mathbf{w})
    \label{eq:llm_coord_prompt}
\end{equation}
The coordination SLM processes this prompt, and its output is parsed by $Parse_\Phi$ to extract the coordinated result, typically enforcing a predefined JSON schema:
\begin{equation}
    (\hat{y}_{LLM}, C_{LLM}, r_{LLM}) = (Parse_\Phi \circ \text{SLM}_{coord} \circ \text{Format})(O, \mathbf{w})
    \label{eq:llm_coord_output}
\end{equation}
This approach allows for potentially more nuanced fusion of agent outputs, leveraging the reasoning capabilities of the coordination SLM.

\subsection{Final Decision Selection Logic ($F_{final}$)}
\label{subsec:decision_logic}

The final decision function, $F_{final}$, synthesizes outputs from the ML agent ($f_{ML}$) and the coordinator ($\Phi$) using a prioritized hierarchy of rules. This logic relies on three predefined components: an ML override condition ($C_{override}$), class-dependent confidence thresholds ($\tau_{coord}(y)$), and a tie-breaking weight ($w_1(y)$).

The thresholds for the coordinated output depend on class rarity:
\begin{itemize}
    \item $\tau_{coord\_rare} = 0.4$: Minimum confidence for rare classes ($\{1, 4\}$).
    \item $\tau_{coord\_common} = 0.5$: Minimum confidence for common classes ($\{2, 3\}$).
\end{itemize}
For ambiguous cases where both the ML and coordinator outputs are moderately confident, the tie-breaking weight $w_1(y)$ also favors rare classes:
\begin{equation}
    w_1(y) = \begin{cases} 0.7 & \text{if } y \in \{1, 4\} \\ 0.5 & \text{if } y \in \{2, 3\} \end{cases}
    \label{eq:weighted_comparison_weight}
\end{equation}

The final prediction $\hat{Y}$ is determined by the following sequence, where the first satisfied rule dictates the outcome:
\begin{equation}
\hat{Y} =
\begin{cases}
    \hat{y}_{ML}, & \text{if } C_{override} \text{ is met} \\
    \hat{y}_{coord}, & \text{if } C_{coord} > \tau_{coord}(\hat{y}_{coord}) \\
    \hat{y}_{ML}, & \text{if } c_{ML} \cdot w_1(y) > C_{coord} \cdot (1 - w_1(y)) \\
    \hat{y}_{coord}, & \text{otherwise}
\end{cases}
\label{eq:final_prediction_logic_revised}
\end{equation}
The final system confidence, $C_{final}$, corresponds to the confidence of whichever source provided the prediction $\hat{Y}$:
\begin{equation}
    C_{final} = c_{ML} \cdot \mathbb{I}(\hat{Y} = \hat{y}_{ML}) + C_{coord} \cdot \mathbb{I}(\hat{Y} = \hat{y}_{coord})
    \label{eq:final_confidence}
\end{equation}
where $\mathbb{I}(\cdot)$ is the indicator function. The final tuple $(\hat{Y}, C_{final})$ is the output of the MARBLE system.

\section{Experiments}
\label{section: experiments}
This section delineates the experimental design employed to evaluate the effectiveness of the proposed accident severity classification framework, which integrates machine learning (ML) models, large language model (LLM) prompting strategies, and a multi-agent system. The evaluation systematically compares our approach against a spectrum of established methodologies, leveraging a structured dataset to assess classification accuracy, robustness, and computational efficiency. By benchmarking against traditional ML models, standalone LLM prompting techniques, and state-of-the-art reasoning paradigms, we aim to demonstrate the framework’s superior predictive capabilities and its suitability for real-time traffic safety applications. The experiments are structured to highlight the contributions of each component—ML baseline, LLM reasoning, and multi-agent synthesis—while providing insights into their combined efficacy. In this research, we aim to answer the following research questions:

\noindent
\hangindent=1.5em \textbf{\circledgreen{1}} How does a multi-agent system with small language models (SLMs) enhance prediction accuracy compared to monolithic ML and prompting approaches for accident severity classification?

\noindent
\hangindent=1.5em \textbf{\circledgreen{2}} How do different coordination strategies (rule-based vs. LLM-based) affect overall system performance under the same underlying conditions?

\noindent
\hangindent=1.5em \textbf{\circledgreen{3}} What is the impact of selectively including or excluding specific domain-specialized SLM agents (e.g., Environmental, Temporal) on the overall accuracy of the MARBLE system, and what insights does this provide regarding the relative importance of the feature subsets utilized by these agents?

\noindent
\hangindent=1.5em \textbf{\circledgreen{4}} How does MARBLE perform under extreme class imbalance, and how do different coordination strategies respond to varying severity label distributions?

\subsection{Experimental Setup}
\label{subsec:experimental_setup}

We evaluate the MARBLE framework against two categories of baselines to benchmark its statistical and reasoning capabilities. The first category includes traditional machine learning: RandomForest and GradientBoosting and LSTM. These models, chosen for their strong performance on structured data, are trained on the complete feature set $\mathcal{F}$ and serve as robust statistical baselines. The second category comprises four standalone LLM prompting strategies designed to isolate reasoning performance without our agentic structure: basic prompting (a minimal baseline), Chain-of-Thought (CoT)~\cite{chain_of_thought_prompting_elicits_reasoning_in_large_language_models}, Self-Consistency with CoT (CoT-SC)~\cite{self_consistency_improves_chain_of_thought_reasoning_in_language_models}, Least-to-Most (L2M)~\cite{least_to_most_prompting_enables_complex_reasoning_in_large_language_models}, Graph-of-Thoughts (GoT) ~\cite{graph_of_thoughts_solving_elaborate_problems_with_large_language_models}, Tree-of-Thought (ToT)~\cite{tree_of_thoughts_deliberate_problem_solving_with_large_language_models} and Chain-of-Draft (CoD) ~\cite{CoD}.

For the MARBLE agents ($a \in \mathcal{A}_{SLM}$), we exclusively selected open-source Small Language Models (SLMs) to ensure the system can run efficiently on consumer-grade hardware and meet the low-latency requirements of real-time deployment. Our evaluation therefore focuses on models in the 1B-3B parameter range, specifically: Gemma-3 1B, LLaMA 3.2 1B, Mistral 1.3B, LLaMA 3.2 3B, and Smollm2 1.7b. By selecting models from distinct architectural families (Gemma, LLaMA, Mistral), we can assess the robustness of the MARBLE framework to the choice of the underlying SLM. This systematic evaluation allows us to measure the impact of SLM selection on key metrics such as prediction accuracy, inference latency, and the quality of generated reasoning ($r_a$).

\begin{figure*}[htbp]
    \centering
    \includegraphics[width=1\linewidth]{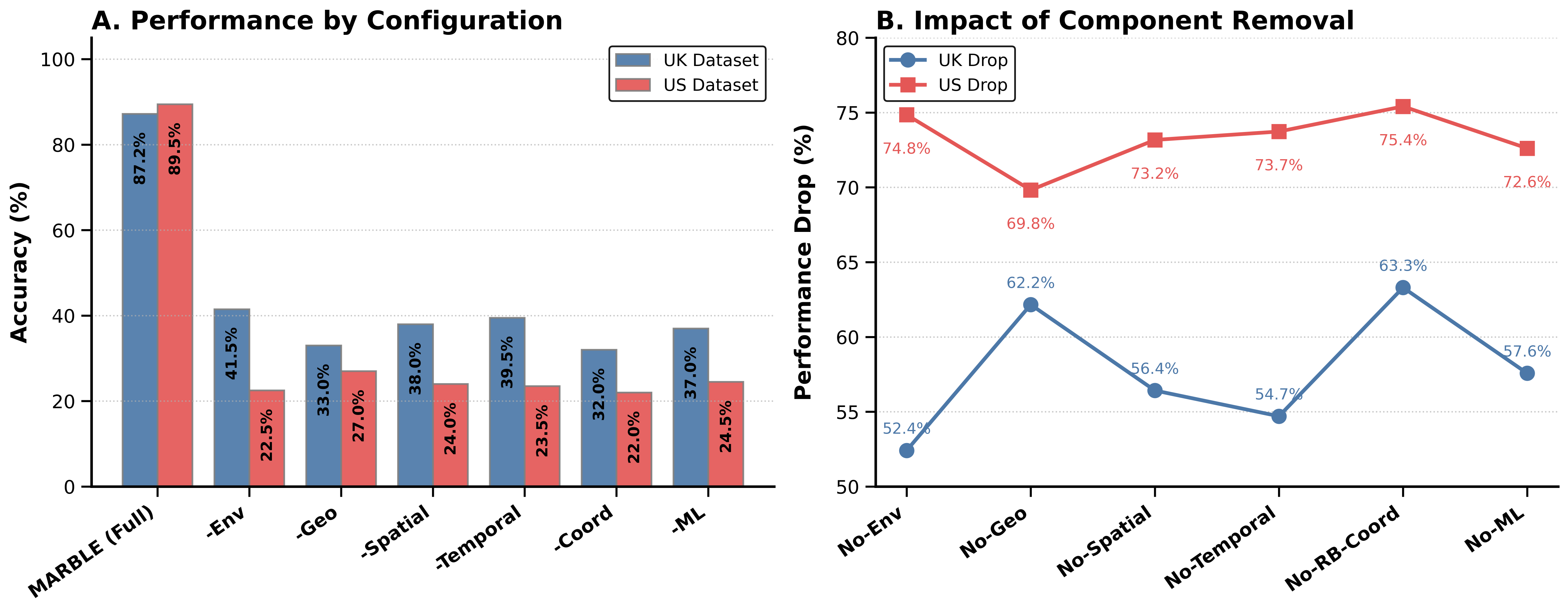}
    \caption{Ablation results on UK and US datasets. Panel A shows accuracy per agent-removed setting; Panel B shows relative accuracy drop.}
    \label{fig:agent-abliation}
\end{figure*}

\section{Results}
\label{sec:results_metrics}

This section elucidates the empirical outcomes of the proposed accident severity classification framework, evaluating the performance of machine learning (ML) models, large language model (LLM) prompting strategies, and the hybrid multi-agent system across UK and US traffic accident datasets. The results, detailed in (Table~\ref{tab:marble_vs_baselines}), compare the efficacy of traditional methods and various LLM SOTA prompting techniques with the innovative Agent-Hybrid approach. Leveraging LLaMA 3.2 3B Instruct executed on small language models (SLMs), the framework demonstrates substantial enhancements in predictive accuracy and robustness, particularly through multi-agent synthesis. These findings affirm the research objectives of improving classification performance and delivering actionable insights into accident severity drivers, with performance assessed using a comprehensive set of evaluation metrics outlined below.

\subsection*{RQ1: Performance Comparison}
Among traditional baselines, tree-based ensembles such as Random Forest and Gradient Boosting achieve moderate performance, with accuracies ranging between 41\% and 45\%, while Support Vector Machines lag slightly behind. Deep learning baselines, exemplified by Long Short-Term Memory (LSTM) models, perform marginally better, peaking at 47.3\% on the US dataset. These outcomes, while consistent with prior literature, are constrained by the limited training data available: each model was trained on only 500 samples per severity class, reflecting a low-resource regime. Such data scarcity significantly impacts ML and DL models, which typically depend on large-scale, balanced corpora to extract generalizable patterns.

\begin{figure*}[htbp]
    \centering
    \includegraphics[width=1\linewidth]{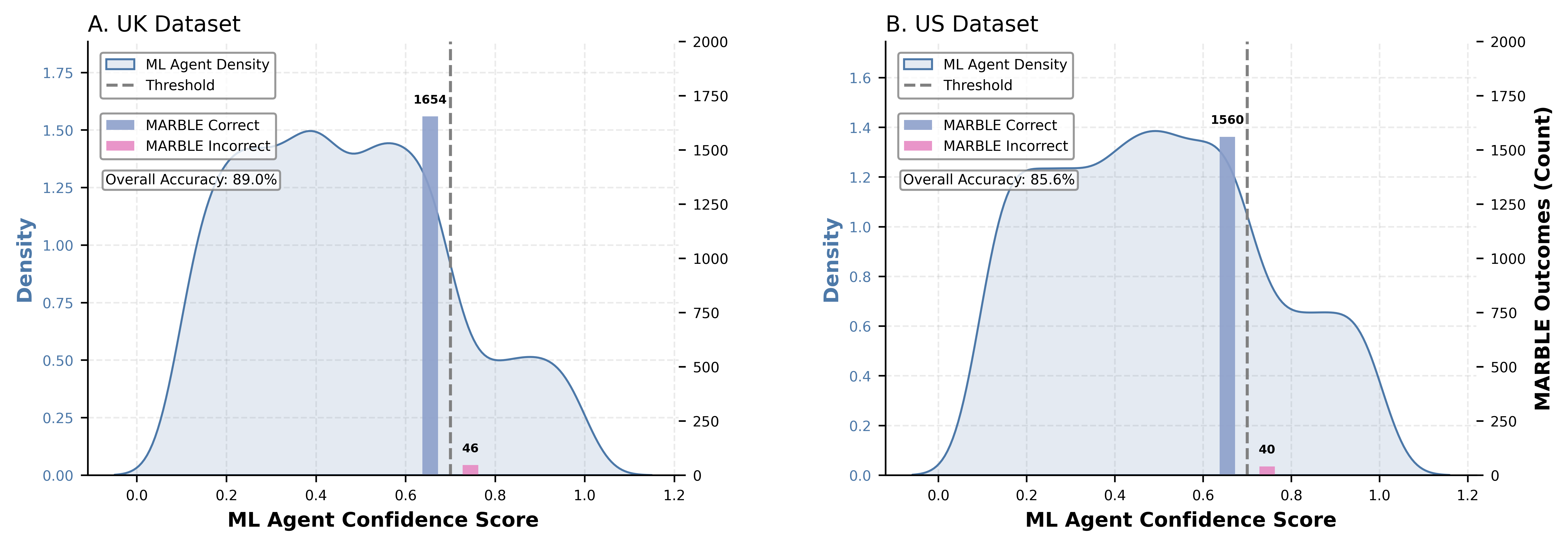}
    \caption{ML agent confidence distributions and MARBLE outcomes on UK (A) and US (B) datasets. Bars show correct and incorrect predictions near the confidence threshold. MARBLE corrects most low-confidence errors, maintaining high accuracy.}
    \label{fig:ML-agent-Confidence}
\end{figure*}

Prompting-based methods, despite employing advanced inference strategies such as Chain-of-Thought, Self-Consistency, Graph-of-Thought, and Tree-of-Thought, yield underwhelming results in this context, with accuracies rarely exceeding 32\%. These results underscore the difficulty of directly applying large language models to structured tabular prediction tasks without explicit schema alignment or specialized tuning. Moreover, the flat prompting strategies employed here—executed via single call inference over entire feature vectors—suffer from context saturation and reasoning entanglement, leading to brittle and inconsistent performance.

In stark contrast, MARBLE achieves a substantial improvement, attaining 89.5\% and 89.8\% accuracy on the UK and US datasets respectively. These gains are accompanied by consistently high precision and recall across all classes. Notably, MARBLE surpasses the best-performing ML baseline by more than 42 percentage points and outperforms all prompting-based approaches by over 58 points. This performance is achieved using a compact SLM model (1.7B parameters) without any large-scale pretraining or dataset-specific tuning, demonstrating that modular agentic reasoning—driven by localized prompts and semantic decomposition—can deliver high-fidelity predictions in low-data, imbalanced scenarios.
\begin{figure}
    \centering
    \includegraphics[width=1\linewidth]{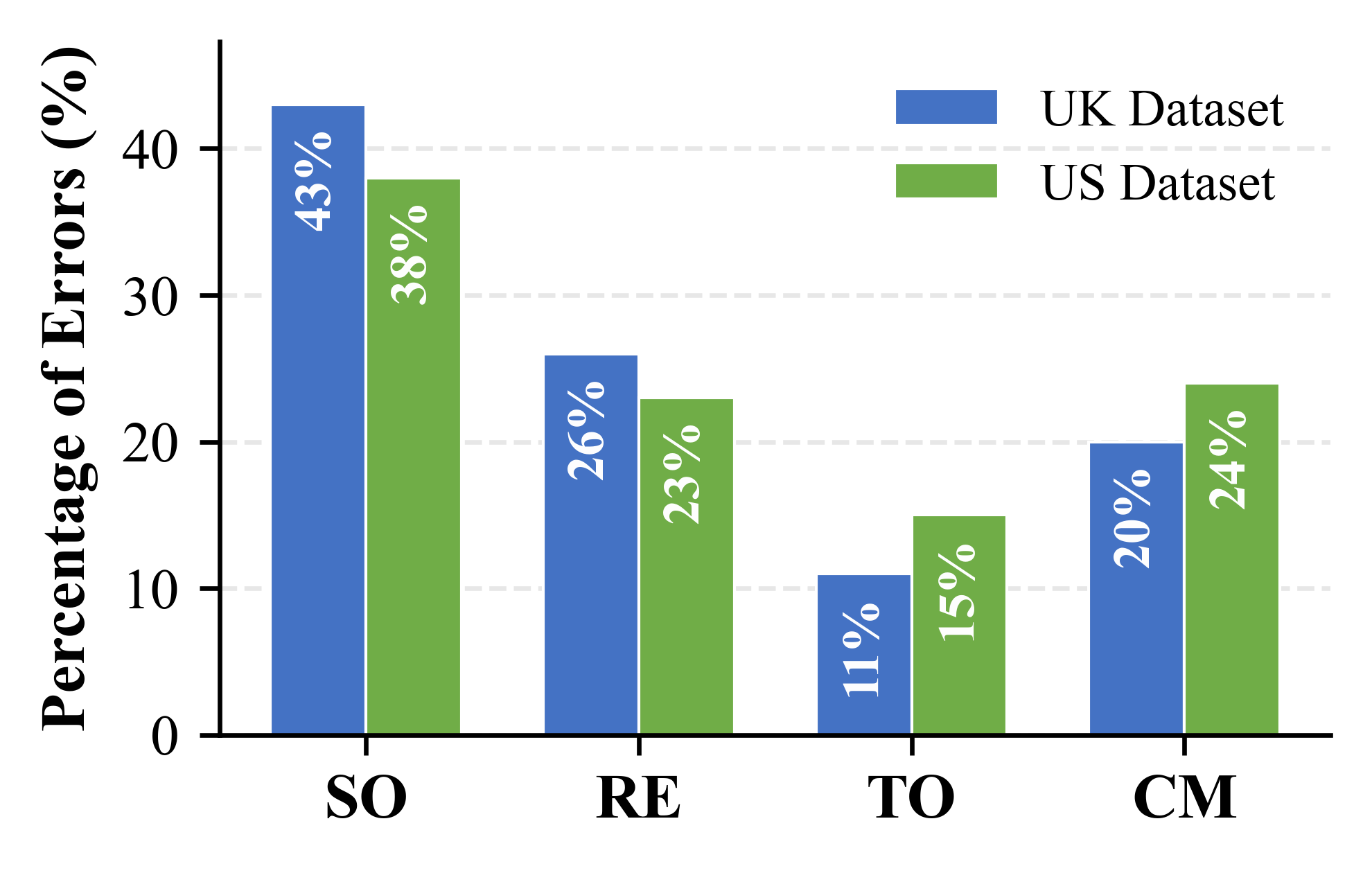}
    \caption{Distribution of failure modes for the MARBLE system, based on analysis \ref{fig:ML-agent-Confidence} (SO = Structured Output, RE = Reasoning Error, TO = Timeout Error, CM = Contextual Misunderstanding)}
\label{fig:error_analysis}
\end{figure}
\subsection*{RQ2: Impact of Coordination Strategy on Performance}

To investigate how coordination strategy influences MARBLE’s effectiveness, we evaluate two distinct approaches mentioned in \ref{sec:methodology} for aggregating agent predictions: lightweight rule-based coordination versus reflective reasoning via small language models (SLMs). Table~\ref{tab:marble_coordination_comparison_final_no_pipe} presents the comparative results across both the UK and US datasets under consistent agent configurations and identical feature inputs.

Within the LLM-based coordination setup, four commonly used open-source SLMs are assessed. Despite using prompting strategies such as Chain-of-Thought and Least-to-Most within the coordination layer, performance across models such as Gemma-3 1B, LLaMA 3.2 1B/3B, and Mistral 1.3B remains moderate—ranging from 65.8\% to 73.4\% in accuracy. These results reflect the inherent challenge of using general-purpose language models to perform real-time judgment fusion over structured, multi-agent outputs, particularly when contextual reasoning must be guided solely through prompt templates.

In contrast, the rule-based coordination variant—where agent outputs are fused via weighted voting and threshold-based consensus—yields significantly higher performance across all model configurations. Notably, accuracy climbs steadily across models and peaks with the Smollm2-1.7B-enhanced MARBLE configuration, which reaches 89.5\% and 89.8\% accuracy on the UK and US datasets respectively. This result outperforms all LLM-based variants by a margin of up to 23.7 percentage points, suggesting that structured aggregation logic, when paired with semantically aligned agents, offers a more stable and computationally efficient inference path.
These findings reinforce our hypothesis that while LLM-based reflection is theoretically expressive, it remains brittle under prompt complexity and input uncertainty. In contrast, rule-based coordination—though simpler—better leverages the modular structure of MARBLE, translating local predictions into high-confidence global decisions without introducing reasoning overhead or interpretability loss.

\begin{figure*}[htbp]
    \centering
    \includegraphics[width=1\linewidth]{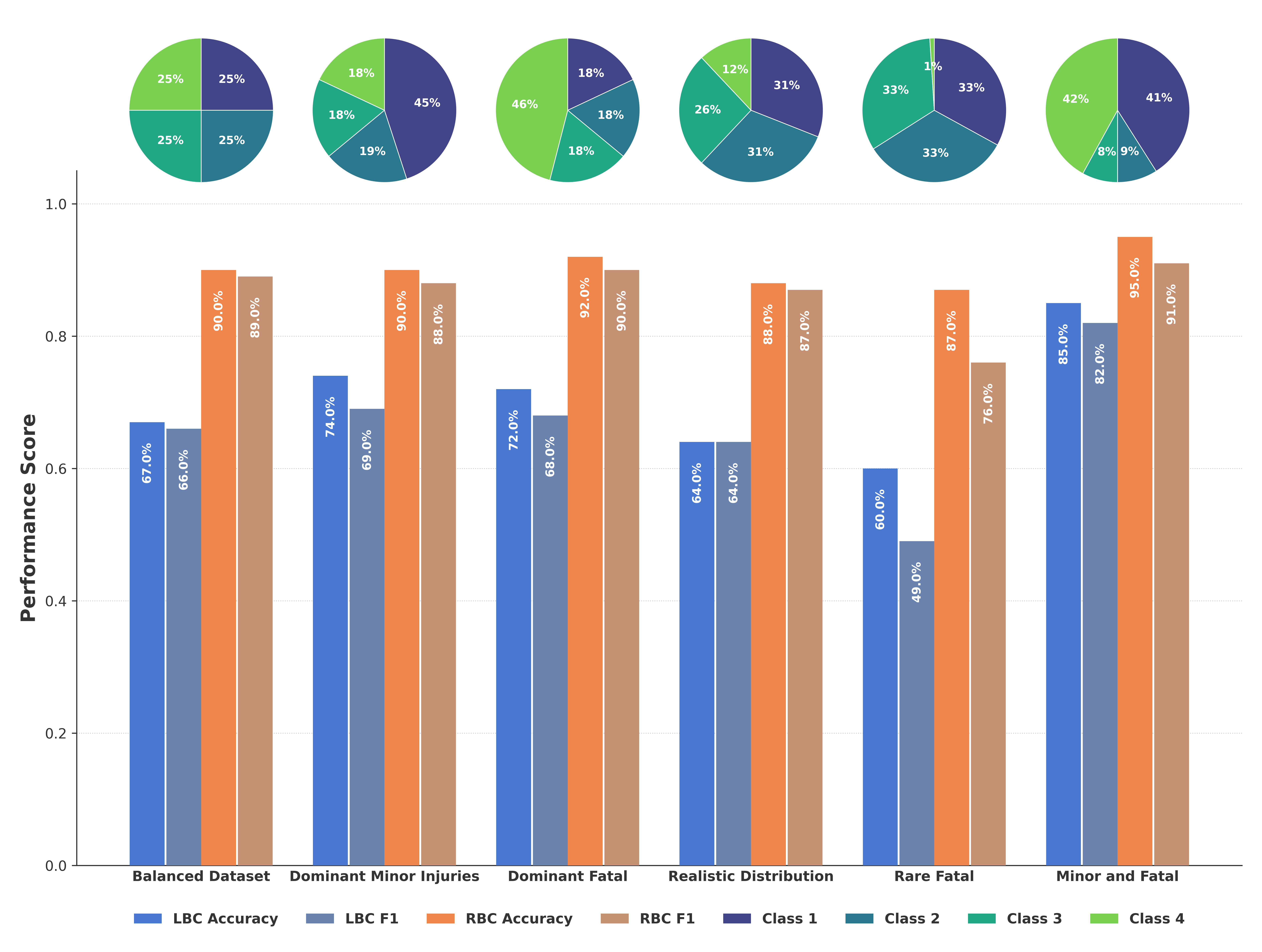}
    \caption{Performance of MARBLE under six simulated class imbalance scenarios in the US dataset. Each setting varies the severity class distribution (top pie charts), with evaluation across RBC and LBC. RBC consistently achieves higher accuracy and F1 scores, especially under extreme skew (e.g., Rare Fatal), highlighting its robustness to label imbalance.}
    \label{fig:robustness-class-blance}
\end{figure*}

\subsection*{RQ3: Agent Importance through Performance Decomposition}

To evaluate the contribution of individual agents to MARBLE’s overall performance, we conduct a systematic ablation study in which each agent is removed independently while the remaining system components are held constant. The results, shown in Figure~\ref{fig:agent-abliation}, span both the UK and US datasets. Panel A reports the absolute classification accuracy for each configuration, whereas Panel B illustrates the corresponding percentage drop relative to the full MARBLE system. The complete MARBLE configuration yields 87.2\% and 89.5\% accuracy on the UK and US datasets, respectively. Excluding the environmental agent results in the most substantial accuracy decline falling to 41.5\% and 22.5\% demonstrating the high predictive value of environment-related features such as weather and lighting. Similar performance degradation is observed when removing the ML Agent, Temporal Agent, or Coordinator, underscoring their pivotal role in sustaining robust cross-agent inference. In contrast, removing the Geo-spatial or Spatial agents leads to smaller, though still measurable, reductions in accuracy.

Panel B further quantifies these effects, with the absence of the Environmental Agent yielding the steepest relative performance drop: 52.4\% on the UK dataset and 74.8\% on the US dataset. The Coordinator Agent also exhibits substantial impact, reaffirming the importance of structured decision fusion in MARBLE’s architecture.

These results highlight the asymmetrical contribution of each agent and validate the architectural decoupling strategy. Unlike monolithic models that assume uniform feature relevance, MARBLE enables localized reasoning and per-domain importance estimation through its modular design. The observed ablation trends confirm that high predictive performance emerges not from any single agent, but from the coordinated interplay between semantically specialized reasoning components and the central coordinator.

\subsection*{RQ4: Robustness to Class Imbalance}
To evaluate MARBLE’s performance under varying degrees of class imbalance, we simulate six distinct label distributions over the US accident dataset—ranging from perfectly balanced to highly skewed toward either minor or fatal injury classes. Each configuration is evaluated under both coordination strategies: Rule-Based Coordination (RBC) and LLM-Based Coordination (LBC). The results, summarized in Figure~\ref{fig:robustness-class-blance}, include accuracy and macro F1 score for each coordination mode, along with the label distribution for each scenario.

Under all imbalance regimes, MARBLE with RBC consistently outperforms its LBC counterpart, maintaining accuracy above 87\% and F1 scores above 85\% even in challenging skew conditions (e.g., Rare Fatal and Minor-Fatal distributions). In contrast, LBC exhibits greater sensitivity to distributional shifts, with performance degrading to 60\% accuracy and 49\% F1 in the Rare Fatal case. This suggests that while LLM-based coordination can match RBC under balanced or mildly skewed conditions, it is less stable when class representation becomes sparse.

Notably, MARBLE maintains its robustness even when rare severity classes dominate the distribution, demonstrating its capacity to generalize without requiring explicit data balancing or synthetic augmentation. The modular agentic architecture, paired with deterministic coordination logic, enables stable reasoning even under severe distributional shifts—whereas prompt-based reflection struggles to resolve inter-class imbalance without confusion or dilution of agent input relevance.

These results affirm MARBLE’s design as both data-efficient and imbalance-resilient, with rule-based coordination offering a more reliable pathway for high-confidence inference in skewed, real-world accident severity distributions.

\begin{figure}
    \centering
    \includegraphics[width=1\linewidth]{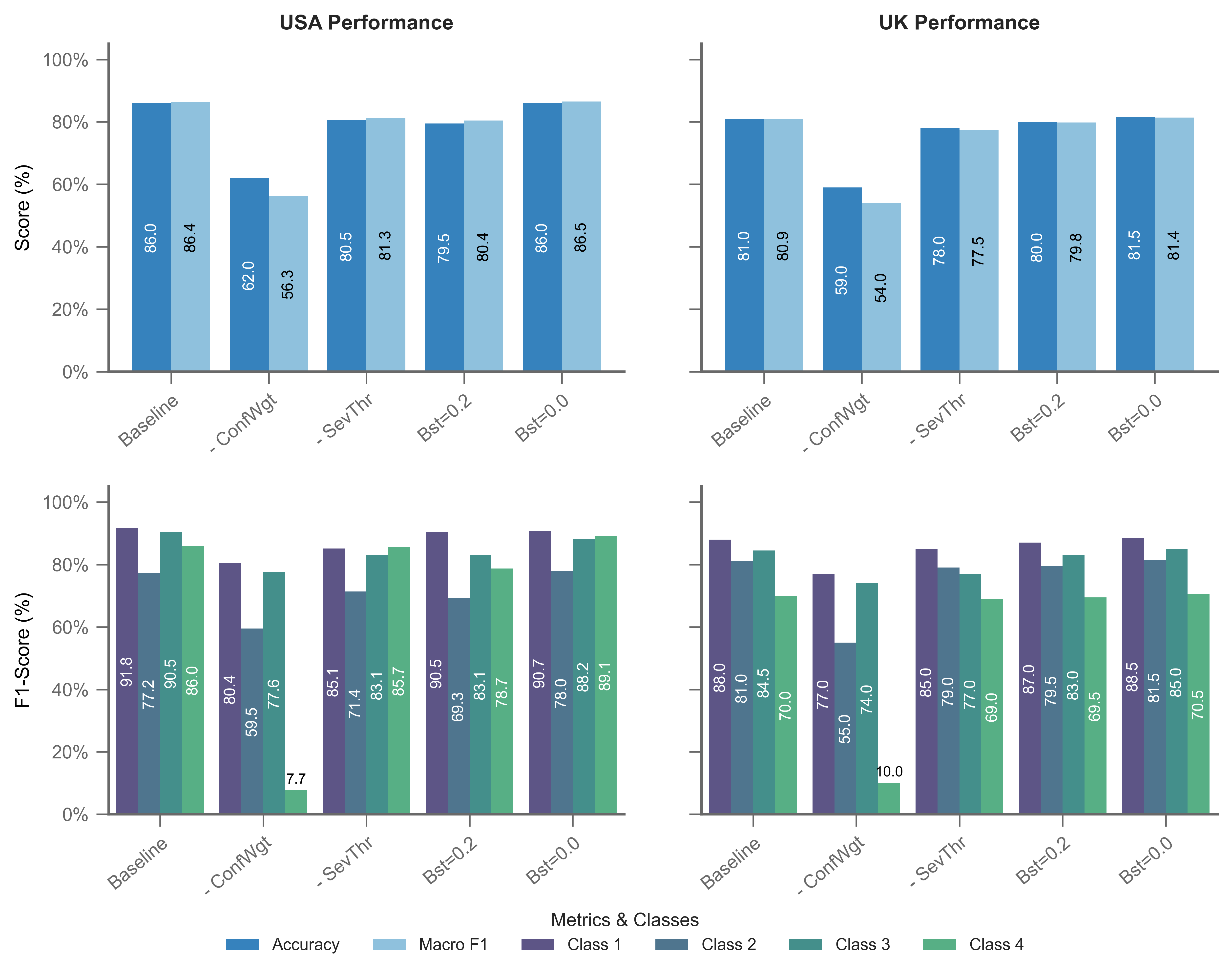}
    \caption{Ablation of MARBLE under coordination variants: removing confidence weights, severity thresholds, and applying different boost settings. Accuracy and F1 scores drop notably when removing key components, confirming their contribution to robust performance on both US and UK datasets.}
    \label{fig:enter-label}
\end{figure}

\section{Related Work}

Accurate prediction of traffic accident severity is pivotal for enhancing road safety and optimizing emergency response systems. Machine learning (ML) approaches have been widely adopted to map structured accident features such as weather, road conditions, and temporal factors to injury outcomes~\cite{Santos2022}. Classical ML models, including logistic regression and decision trees, demonstrated early success, while ensemble methods and deep learning have more recently offered improved predictive performance~\cite{Santos2022}. To address the demand for model transparency in safety-critical domains, researchers have introduced explainable frameworks. Yang~\textit{et al.}~\cite{Yang2022} proposed a multi-task deep learning framework enhancing interpretability in severity prediction, and Cicek~\textit{et al.}~\cite{Cicek2023} compared multiple explainable models, highlighting competitive performance across techniques. Hybrid architectures such as RFCNN~\cite{Manzoor2021} have also emerged, combining random forest and convolutional neural networks to improve classification accuracy. Nevertheless, challenges persist, particularly the impact of class imbalance and the difficulty in achieving robust generalization across heterogeneous datasets~\cite{Santos2022}.

The structured nature of accident data introduces additional complexities. Although deep learning has achieved remarkable advances in unstructured domains, its application to tabular data remains limited. As Borisov~\textit{et al.}~\cite{Borisov2022} detailed, deep neural networks often underperform compared to tree-based models when handling heterogeneous, small-to-moderate structured datasets. The necessity for extensive preprocessing, such as categorical embeddings and normalization, further complicates model development~\cite{ShwartzZiv2021}. Moreover, the interpretability gap remains significant; while black-box models achieve high expressiveness, they offer limited diagnostic insight~\cite{Santos2022}. These challenges underline the limitations of relying solely on conventional deep learning architectures for accident severity prediction and emphasize the need for frameworks that balance predictive power with interpretability.

Agentic AI offers an alternative paradigm by structuring intelligence as cooperation among specialized autonomous agents. Within transportation research, multi-agent systems have been employed to model interactions among vehicles, infrastructure, and human behavior, effectively capturing emergent phenomena that single-agent models overlook~\cite{Li2024}. Multi-agent reinforcement learning (MARL) further extends this by enabling decentralized decision-making in dynamic environments~\cite{Li2024}. Recent innovations integrate language models within agentic systems, facilitating communication and collaboration among agents through natural language~\cite{Li2024}. This evolution positions agentic architectures as a compelling framework for structured prediction tasks, such as accident severity analysis, where multiple interdependent factors influence outcomes. However, deploying full-scale large language models (LLMs) as agents is impractical for many applications due to computational and latency constraints, necessitating more efficient alternatives.

Recent advances in small language models (SLMs) and prompting techniques provide promising solutions for scalable agentic systems. Prompting strategies enable models to adapt to new tasks without exhaustive retraining by structuring inputs to elicit the desired reasoning behavior~\cite{Liu2023}. Brown~\textit{et al.}~\cite{Brown2020} demonstrated the potential of few-shot prompting in LLMs, and subsequent work, such as chain-of-thought prompting~\cite{Wei2022}, has shown that guiding models to reason step-by-step significantly enhances their problem-solving capabilities. Techniques like knowledge distillation and instruction-tuning further enable smaller models to inherit reasoning patterns from larger counterparts~\cite{Wei2022,Liu2023}. These methods allow for the construction of lightweight, specialized agents capable of high-quality inference while maintaining efficiency.

Integrating these strands, the proposed MARBLE system seeks to overcome the limitations identified in prior work. MARBLE adopts a hybrid architecture wherein structured ML models are complemented by a distributed set of domain-specialized SLM-based agents. Each agent focuses on a coherent subset of accident features and applies advanced prompting strategies to perform focused reasoning. A structured coordination mechanism synthesizes agent outputs, ensuring robust, interpretable predictions even under conditions of data incompleteness and severe class imbalance. By uniting structured data modeling, agentic collaboration, and efficient language-based reasoning, MARBLE addresses critical gaps in existing accident severity prediction frameworks.

\section{Limitations}
\label{sec:limitations}
The proposed MARBLE framework, while effective, presents certain limitations inherent to its design. Firstly, the performance of language model-based agents is dependent on the stability and instruction-following capabilities of the chosen SLMs, particularly with complex prompts or diverse inputs, which may necessitate careful prompt engineering and output validation.

Secondly, the coordination mechanism, especially the rule-based variant, assumes consistent adherence to predefined output structures by all agents. Potential deviations in agent outputs could influence the efficacy of the aggregation process, highlighting the importance of robust agent behavior and communication protocols.
Thirdly, the multi-agent inference pipeline naturally incurs greater computational latency compared to single-pass monolithic models. While SLMs offer efficiency gains over larger models, this inherent trade-off may be relevant for applications with stringent real-time constraints.

Finally, this work focused on a centralized coordination architecture. The exploration of alternative multi-agent system topologies (e.g., decentralized, hierarchical) presents opportunities for future research, potentially offering different balances between performance, robustness, and complexity. These aspects represent areas for continued investigation and refinement of the MARBLE approach.

\section{Conclusion}
\label{sec:conclusion}
This paper introduced MARBLE, a hybrid multi-agent framework that synergizes machine learning agent with specialized Small Language Model (SLM) agents to predict accident severity. By decomposing the problem across domain-focused agents using a structured, rule-based coordination strategy, MARBLE overcomes the limitations of monolithic models and direct large-language-model prompting. On UK and US datasets, MARBLE achieved nearly 90\% accuracy, decisively outperforming traditional classifiers and advanced prompting methods that plateaued below 48\%. This performance underscores the framework's robustness and its ability to handle severe class imbalance.

MARBLE demonstrates that hybrid multi-agent systems using efficient SLMs can achieve high-fidelity, interpretable predictions on complex structured data. This represents a promising paradigm for safety-critical applications. Future work may explore adaptive coordination mechanisms and alternative agent interaction topologies.

\appendix
\subsection*{Dataset Descriptions}
\label{app:datasets}
The analyses in this paper were conducted on two prominent public accident datasets from the United States and the United Kingdom. While the SLM agents in MARBLE process each accident as an independent reasoning task, the scale and diversity of these datasets are essential for the robust evaluation of the system's overall performance and for training its integrated machine learning component. The US data was derived from the countrywide dataset introduced by Moosavi et al. \cite{moosavi2019countrywidetrafficaccidentdataset}, from which we extracted a subset of 150,000 instances after pre-processing. Similarly, the UK data was sourced from the STATS19 database, a public repository from the UK Department for Transport \cite{Tsiaras2019UKRoadSafety}. Our analysis used a curated subset of these records that yielded 180,000 instances following a similar filtering and cleaning pipeline.

\subsection*{SLM Configuration}
The domain-specific SLM agents utilized models (bfloat16 precision). The decoding strategy was optimized for factual reasoning with the following hyperparameters: a temperature of 0.2, top-p sampling set to 0.90, a repetition penalty of 1.1, and a maximum output of 256 new tokens (set for speed and structure output). To ensure system stability, an 8-second timeout guardrail was enforced. The primary prompting strategy for the agents was Chain-of-Thought (CoT), with Least-to-Most (L2M) used for the comparative LLM-based Coordinator agent.

\begin{table}[h!]
\centering
\caption{Key features allocated to each domain-specific agent. The full feature set includes additional variables derived from the source datasets.}
\label{tab:features_list}
\begin{tabular}{@{}l l l@{}}
\toprule
\textbf{Agent Domain} & \multicolumn{2}{c}{\textbf{Associated Features}} \\
\cmidrule(l){2-3} % a partial rule under the "Associated Features" header
% --- Environmental Section ---
\multirow{3}{*}{\textbf{Environmental}} & Weather Conditions & Temperature \\
 & Light Conditions & Wind Speed \\
 & Visibility & Humidity \\
\addlinespace % Adds a little space before the next group
% --- Temporal Section ---
\multirow{3}{*}{\textbf{Temporal}} & Day of Week & Weekend/Holiday \\
 & Time of Day & Day of Year \\
 & Month & Part of Day \\
\addlinespace
% --- Infrastructural Section ---
\multirow{3}{*}{\textbf{Infrastructural}} & Road Type & Road Surface \\
 & Junction Detail & Special Conditions \\
 & Speed Limit & Carriageway Hazards \\
\addlinespace
% --- Spatial/Dynamic Section ---
\multirow{3}{*}{\textbf{Spatial/Dynamic}} & Point of Impact & Longitude \\
 & Travel Distance & Latitude \\
 & Vehicle Manoeuvres & Spatial Extent \\
\bottomrule
\end{tabular}
\end{table}

\bibliographystyle{IEEEtran}
\bibliography{references}
\end{document}